\documentclass{article}
\usepackage[utf8]{inputenc}
\usepackage{microtype}
\usepackage{graphicx}
\usepackage{subcaption}
\usepackage{booktabs} 
\usepackage{hyperref}

\usepackage[preprint]{icml2026}
\usepackage{amsmath}
\usepackage{amssymb}

\newcommand{\std}[1]{{\scriptsize $\pm$#1}}
\usepackage{mathtools}
\usepackage{amsthm}
\usepackage{comment}
\usepackage{tabularx}
\usepackage{ragged2e}
\usepackage{bbm}
\usepackage{multirow}
\usepackage{algorithm, algorithmic}
\usepackage[table]{xcolor}
\definecolor{modelgray}{gray}{0.93}
\usepackage[most]{tcolorbox}
\usepackage{enumitem}
\usepackage{verbatim}
\usepackage[capitalize,noabbrev]{cleveref}
\usepackage{listings}
\theoremstyle{plain}

\theoremstyle{definition}

\theoremstyle{remark}

\usepackage[textsize=tiny]{todonotes}

\lstdefinestyle{jsonbox}{
  basicstyle=\ttfamily\footnotesize,
  breaklines=true,
  breakatwhitespace=false,
  columns=fullflexible,
  showstringspaces=false,
  keepspaces=true,
}
\newtcolorbox{promptbox}[2][]{
  enhanced,
  colback=blue!3,
  colframe=blue!45!black,
  coltitle=white,
  colbacktitle=blue!55!black,
  title={#2},
  fonttitle=\bfseries\small,
  fontupper=\small,
  sharp corners,
  boxrule=0.5pt,
  #1
}

\icmltitlerunning{Submission and Formatting Instructions for ICML 2026}

\begin{document}

\twocolumn[
  \icmltitle{Mitigating Cognitive Inertia in Large Reasoning Models via Latent Spike Steering}



  \icmlsetsymbol{equal}{*}

  \begin{icmlauthorlist}
    \icmlauthor{Seojin Lee}{yyy}
    \icmlauthor{ByeongJeong Kim}{yyy}
    \icmlauthor{Hwanhee Lee}{yyy}
  \end{icmlauthorlist}

  \icmlaffiliation{yyy}{Department of Artificial Intelligence, Chung-Ang University, Seoul, Korea}

  \icmlcorrespondingauthor{Hwanhee Lee}{hwanheelee@cau.ac.kr}

  \icmlkeywords{Large Reasoning Models, Overthinking, Rigidity}
  \vskip 0.3in
]

\printAffiliationsAndNotice{}  

\begin{abstract}
While Large Reasoning Models (LRMs) have achieved remarkable performance by scaling test-time compute, they frequently suffer from \textbf{Cognitive Inertia}, a failure pattern manifesting as either \textit{overthinking} (inertia of motion) or \textit{reasoning rigidity} (inertia of direction). Existing detection methods, typically relying on superficial textual heuristics like self-correction tokens, often fail to capture the model's unvoiced internal conflicts. To address this, we propose \textbf{STARS} (Spike-Triggered Adaptive Reasoning Steering), a training-free framework designed to rectify cognitive inertia by monitoring latent dynamics. STARS identifies \textbf{Cognitive Pivots}—critical moments of reasoning transition—by detecting distinct $L_2$ distance spikes in the hidden states. Upon detection, the framework employs geometric trajectory analysis to diagnose the structural nature of the transition and injects state-aware language cues to steer the model in real-time. Our experiments across diverse benchmarks confirm that STARS efficiently curtails redundant loops while improving accuracy through the adaptive correction of erroneous trajectories. STARS offers a robust, unsupervised mechanism to optimize the reasoning process of LRMs without requiring additional fine-tuning.
\end{abstract}

\begin{figure}[t]
    \centering
    \includegraphics[width=\linewidth]{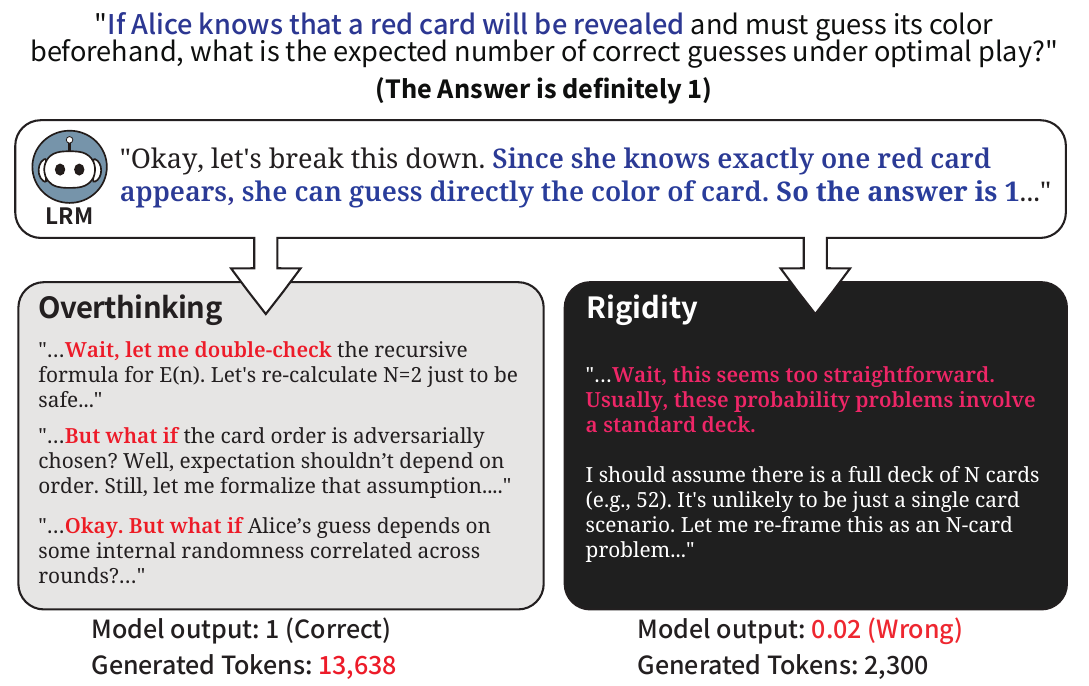}
    \caption{\textbf{Cognitive Inertia in LLMs.} 
    An illustration of two failure modes—\emph{overthinking} (inertia of motion) and \emph{reasoning rigidity} (inertia of direction).}
    \label{fig:intro_inertia}
    \vspace{-7mm}
\end{figure}

\section{Introduction}

The paradigm of Large Language Models (LLMs) has recently shifted from parameter scaling to scaling test-time compute.
Models like o1~\citep{jaech2024openai} and DeepSeek-R1~\citep{guo2025deepseek} show that extending reasoning chains can yield performance gains comparable to massive parameter scaling~\citep{muennighoff2025s1}.
By decomposing complex problems into intermediate steps, these Large Reasoning Models (LRMs) have achieved state-of-the-art results in mathematical and logical domains.

However, this reliance on extended generation has introduced new forms of inference failure, which we collectively term \textbf{Cognitive Inertia}. 
As illustrated in Figure~\ref{fig:intro_inertia}, this phenomenon manifests in two distinct failure modes that mirror the properties of physical inertia. 
First, unguided scaling often leads to \textit{overthinking} (inertia of motion), where models spiral into redundant verification loops for simple queries. 
For instance, on the left of Figure~\ref{fig:intro_inertia}, the model generates over 13,000 tokens by repeatedly double-checking recursive formulas and adversarial assumptions, only to reach a trivial solution without any real quality gain~\citep{chen2025do,ghosal2025does}. 
Second, models frequently exhibit \textit{reasoning rigidity} (inertia of direction), where strong parametric priors stubbornly override explicit user constraints. 
As seen in the right part of the figure, even when told only a single card exists, the model's ``directional inertia'' forces it back to a familiar 52-card deck template, leading to a confidently incorrect output~\citep{wu2024reasoning, jang2025reasoning}.
These failures suggest that simply allowing a model to ``think longer'' does not guarantee better outcomes; rather, it necessitates mechanisms to actively monitor and regulate the reasoning trajectory.

To mitigate these inefficiencies, recent works on LRMs have proposed various control mechanisms. 
Prompt-based approaches~\citep{han2025token} attempt to enforce conciseness, yet are often bypassed by the model's internal biases. 
More dynamic approaches, such as DEER~\citep{yang2025dynamic} and ConCISE~\citep{qiao2025concise}, introduce early-exit strategies based on token-level entropy or confidence to truncate unnecessary generation. 
Despite their utility in curbing simple redundancy, we find that these surface-level interventions suffer from two critical shortcomings. 
First, token-level metrics often fail to capture the model's unvoiced internal conflicts, making them unreliable proxies for reasoning quality~\citep{wang2025wait}.
Second, and more critically, early-exit mechanisms are fundamentally \textit{suppressive}—they can stop a model from overthinking, but they cannot \textit{steer} a model that is rigidly pursuing an incorrect trajectory. 
In scenarios involving reasoning rigidity, the model requires active redirection rather than mere cessation.

To address the limitations of surface-level monitoring in regulating cognitive inertia, we propose a shift toward analyzing the latent dynamics of LRMs. 
We discover that the signals required for such regulation are rooted in specific functional boundaries we term \textbf{Cognitive Pivots}, where the model undergoes a significant reconfiguration in its reasoning mode. 
We find that these pivotal moments can be effectively identified through \textbf{Hidden State Spikes}, which manifest as sudden, high-magnitude reconfigurations in the hidden states. 
But this latent spike is fundamentally neutral, representing a state of elevated plasticity rather than an inherent indicator of error. 
To differentiate between these distinct outcomes, we conceptually distinguish between routine \textbf{Functional Pivots} and pathological \textbf{Critical Pivots}. 
This taxonomy establishes a systematic basis for diagnosing internal conflicts and determining precisely when real-time intervention is required to rectify reasoning trajectories.

Building on this foundation, we introduce \textbf{STARS (Spike-Triggered Adaptive Reasoning Steering)}, a training-free framework that employs a three-stage cycle: \textbf{Detect}, \textbf{Select}, and \textbf{Steer}. 
STARS first detects latent spikes in real-time, then diagnoses whether they indicate \textbf{Critical Pivots} using geometric analysis to resolve signal neutrality. 
Finally, the framework steers the reasoning path by injecting state-aware language cues: a \textbf{Shifting Suffix} validates and grounds legitimate reasoning transitions to overcome rigid parametric priors, while a \textbf{Loop Breaker Suffix} disrupts redundant cycles by prompting the model to terminate unproductive ``inertia of motion'' and consolidate its progress toward a final answer.
This closed-loop approach allows LRMs to autonomously navigate complex trajectories without the need for additional fine-tuning.

We evaluate STARS on diverse reasoning benchmarks, including instability-oriented diagnostics~\citep{jang2025reasoning,wu2024reasoning} and high-complexity problems such as AIME~\citep{aime}.
Across both settings, STARS consistently improves performance while reducing unnecessary token usage. 
Rather than uniformly truncating reasoning, STARS adaptively allocates computation when beneficial, guiding models toward correct reasoning trajectories without sacrificing accuracy. Overall, these results indicate that spike-triggered steering provides a robust, unsupervised mechanism to optimize the reasoning process of LRMs.

\begin{figure*}[h]
    \centering
    \begin{subfigure}[b]{0.48\textwidth}
        \centering
        \includegraphics[width=\linewidth]{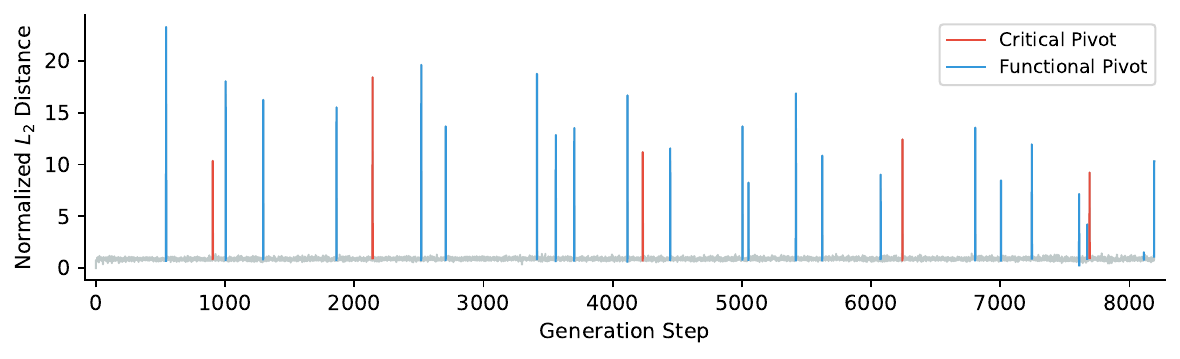}
        \caption{\textbf{DeepSeek-7B} (Layer 13)}
        \label{fig:vis_7b}
    \end{subfigure}
    \hfill
    \begin{subfigure}[b]{0.48\textwidth}
        \centering
        \includegraphics[width=\linewidth]{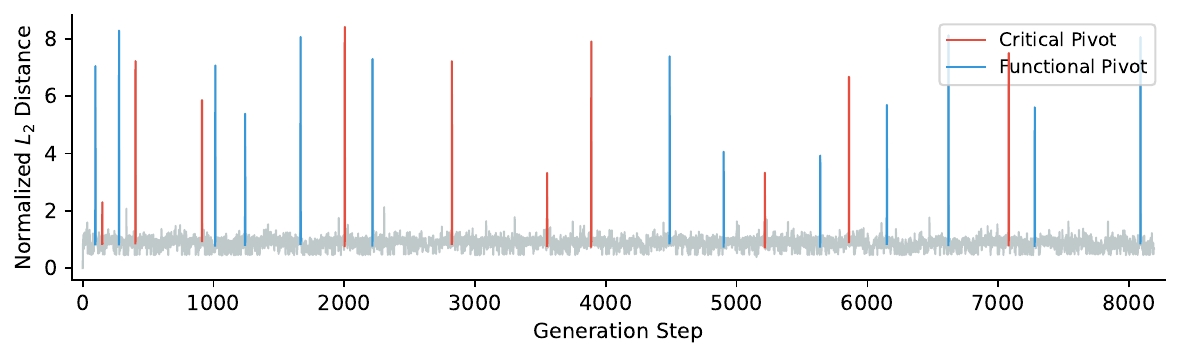}
        \caption{\textbf{Qwen3-4B} (Layer 2)}
        \label{fig:vis_4b}
    \end{subfigure}
    \caption{\textbf{Visualization of latent $L_2$ spike patterns across different models.}
    We plot high-magnitude latent displacement events (spikes), measured by the layer-wise $L_2$ distance between consecutive hidden states, for two distinct models processing the same problem instance.
    Despite exhibiting similarly prominent spike signals, the two models differ substantially in the distribution and relative prevalence of \textcolor{cyan}{\textbf{Functional}} and \textcolor{red}{\textbf{Critical}} pivots, reflecting heterogeneity in their latent reasoning dynamics across architectures.}
    \vspace{-5mm}
    \label{fig:vis_spike}
\end{figure*}

\section{The Hidden State Spike: A Window into Cognitive Inertia}
\label{sec:foundation}
In this work, we formalize \textbf{Cognitive Inertia}, a phenomenon where the generative momentum of a LRM becomes decoupled from its underlying reasoning objective. 
As models scale test-time compute through extended chains of thought, their internal states can develop a structural persistence—a ``cognitive momentum''—that resists mid-course corrections, even when the context provides clear evidence for a pivot or termination. 
We define this inertia as a failure of adaptive control, where the model's internal trajectory becomes self-reinforcing and increasingly immune to both explicit user constraints and its own internal self-correction mechanisms.
To provide a systematic framework for analyzing this failure, we categorize cognitive inertia into two distinct modes based on physical analogies:

\begin{itemize}[leftmargin=*,itemsep=0.1em]
    \item \textbf{Overthinking (Inertia of Motion):} Similar to an object that resists braking, the model persists in redundant verification loops long after reaching a plausible solution. 
    This represents a failure to recognize the sufficiency of existing progress, where generation continues despite diminishing returns or logical completion~\citep{ghosal2025does}.
    
    \item \textbf{Reasoning Rigidity (Inertia of Direction):}
    Similar to an object that resists changing course, the model stubbornly adheres to dominant parametric priors despite explicit instructions to deviate. 
    This \textit{inertia of direction} causes the model to override task-specific constraints—such as counterfactual premises—in favor of familiar, canonical templates~\citep{jang2025reasoning}.
\end{itemize}

\subsection{Cognitive Pivots: The Locus of Reasoning Transition}
\label{sec:cognitive_pivots}
Despite their differing outcomes, these failures share a common temporal structure: they are not continuous, but tend to emerge around specific transition points we term \textbf{Cognitive Pivots}. 
These pivots correspond to functional boundaries where the model’s reasoning mode is in flux (e.g., shifting from calculation to verification).
We hypothesize that at these boundaries, the model's trajectory becomes highly plastic and particularly susceptible to inertial drift. 

While prior works rely on surface-level textual cues to monitor these transitions, such signals offer only an indirect view of the underlying dynamics. 
This motivates a shift toward examining the model's internal representation space as a more faithful diagnostic window.

\subsection{The Hidden State Spike Phenomenon}
\label{sec:spike_phenomenon}
To quantify the internal dynamics of reasoning, we monitor the layer-wise Euclidean distance of the hidden states in the decoder. 
We define the \textbf{Latent Displacement} $\delta^{(l)}_t$ at time step $t$ and layer $l$ as the magnitude of the state update:
\begin{equation}
    \delta^{(l)}_t = \|h^{(l)}_t - h^{(l)}_{t-1}\|_2
\end{equation}
Figure~\ref{fig:vis_spike} illustrates that $\delta^{(l)}_t$ exhibits distinct spike patterns that are clearly distinguishable from background generation noise across layers.
We term these events \textbf{Hidden State Spikes}.
These high-magnitude deviations suggest that the model undergoes substantial internal reconfigurations at specific intervals.
We hypothesize that these spikes characterize transitions in the reasoning process, marking moments where the internal state is highly plastic and potentially vulnerable to the onset of cognitive inertia. 
As visualized in Figure~\ref{fig:pivot_analysis}, these latent fluctuations consistently align with the functional boundaries where the model's reasoning mode is in flux, providing a robust indicator for real-time monitoring well before failures manifest in the generated text.

\begin{figure}[h]
\centering
\footnotesize
\begin{promptbox}{Spike-Aligned Context Windows}
\textbf{Critical Pivot} \\[0.3em]
\texttt{
has a typo and meant that B and D are distinct, but we have to think about the process? \\
Wait, maybe not. Wait, wait! Wait a second. \\
Maybe the problem is in the \textcolor{red}{\textbf{style}}
}
\vspace{5mm}

\textbf{Functional Pivot} \\[0.3em]
\texttt{
$(2 + 8 + 1)/4 = 1025/4$, correct. Oh, nice! \\
So the factors are integers in the \textcolor{red}{\textbf{numer}}
}
\end{promptbox}
\caption{\textbf{Spike-aligned context windows.}
Red-highlighted tokens indicate detected latent spikes.
Each window shows a short preceding context for interpretability.}
\label{fig:pivot_analysis}
\vspace{-7mm}
\end{figure}

\subsection{Semantic Taxonomy: Functional vs. Critical Pivots}
\label{sec:semantic_analysis}
To better understand the role of high-magnitude latent displacements, we examine their associated semantic contexts within the reasoning chain.
As illustrated in Figure~\ref{fig:pivot_analysis}, latent spikes tend to coincide with pivotal turning points in the chain-of-thought.
However, a key observation is that a latent spike is fundamentally \textbf{neutral}: it does not inherently indicate an error, but rather marks a moment of \textbf{elevated plasticity} during which the model undergoes a substantial internal transition.

In typical reasoning trajectories, these transitions most often correspond to \textbf{Functional Pivots}—benign operational shifts required for problem-solving, such as initiating a calculation or switching representational modes. 
In these cases, the spike reflects a necessary reconfiguration, and the surrounding context remains coherent and goal-directed. 
By contrast, certain spikes are associated with textual patterns characteristic of cognitive inertia, including hesitation, redundant self-correction, or repetitive reconsideration; we refer to these as \textbf{Critical Pivots}.

To quantify this taxonomy, we conducted a manual analysis on 151 identified transitions (detailed in Appendix~\ref{sec:appendixC}). 
We found that the majority of high-magnitude latent spikes (\textbf{72.8\%}) correspond to \textbf{Functional Pivots}, while a significant minority (\textbf{27.2\%}) coincide with \textbf{Critical Pivots} that exhibit clear symptoms of cognitive inertia. 
Importantly, this distinction does not arise from the spike token itself, but from the broader trajectory in which the spike is embedded.
These observations underscore the inherent semantic ambiguity of latent spikes when viewed through surface-level textual analysis. 
While spikes are reliable indicators of reasoning transitions, local textual cues often prove insufficient to distinguish between functional and critical pivots in real-time.


\begin{figure*}[!ht]
    \centering
    \includegraphics[width=0.92\linewidth]{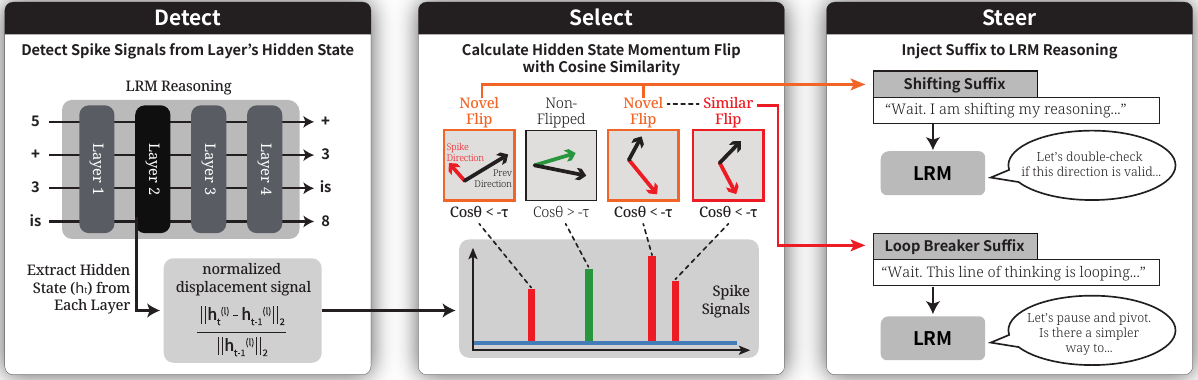}
    \vspace{-1mm}
    \caption{\textbf{Overview of Spike-Triggered Adaptive Reasoning Steering (STARS).}}
    \vspace{-6mm}
    \label{fig:framework}
\end{figure*}


Much like prior token-centric heuristics, textual monitoring remains fundamentally reactive, identifying failures only after the reasoning trajectory has already stabilized into an unproductive loop or diverged beyond simple correction. 
Consequently, relying exclusively on text-dependent interpretation fails to capture the proactive \emph{window of opportunity} offered by these internal reconfigurations. 
This necessitates a framework that treats latent spikes as \textbf{internal transition signals} and employs a more robust, latent-based diagnostic approach to resolve their functional meaning.

\section{STARS: Spike-Triggered Adaptive Reasoning Steering}
\label{sec:method}
We propose \textbf{STARS (Spike-Triggered Adaptive Reasoning Steering)}, a training-free framework designed to counteract cognitive inertia in real-time. 
Building on the insights established in Section~\ref{sec:foundation}, STARS operationalizes latent monitoring into an actionable three-stage cycle as described in Figure~\ref{fig:framework}: (i) \textbf{Detect}, which identifies candidate pivots via normalized spikes; (ii) \textbf{Diagnose}, which analyzes trajectory geometry to resolve signal neutrality; and (iii) \textbf{Steer}, which injects state-aware natural language cues to recalibrate the reasoning path.

\subsection{Stage 1: Detect (Candidate Spike Identification)}
\label{sec:phase_detect}
The first stage of STARS involves isolating meaningful reasoning reconfigurations from stochastic background noise. 
While Section~\ref{sec:foundation} establishes the existence of latent spikes, transforming them into a reliable trigger for intervention requires overcoming the inherent noise in latent dynamics and cross-architecture activation variance. 
We formalize this process through a robust detection procedure designed to identify the precise onset of reasoning transitions.

\paragraph{Robust Spike Signal Extraction} 
To ensure the framework is architecture-aware and scale-invariant, we monitor the normalized displacement signal $\mathcal{S}^{(l)}_t$ for each layer $l$ at time step $t$:
\begin{equation}
\label{eq:normalized_signal}
\mathcal{S}^{(l)}_t = \frac{\|h^{(l)}_t - h^{(l)}_{t-1}\|_2}{\|h^{(l)}_{t-1}\|_2},
\end{equation}
where $h^{(l)}_t$ denotes the hidden state at layer $l$.

Unlike naive thresholding, which is sensitive to outliers, we estimate an adaptive threshold $\tau^{(l)}$ using the \textbf{Median Absolute Deviation (MAD)}. 
This robust statistical measure allows us to define the set of significant pivot events $\mathcal{P}^{(l)}$ as those exceeding the baseline volatility:
\begin{equation}
\tau^{(l)} = \operatorname{median}(\mathcal{S}^{(l)}) + k \cdot \operatorname{MAD}(\mathcal{S}^{(l)}),
\end{equation}
where $k$ is a sensitivity parameter calibrated to maximize the detection of latent reconfigurations.
We then isolate the significant reasoning transitions by defining the set of pivot events $\mathcal{P}^{(l)}$ as the collection of signals exceeding this robust threshold: 
\begin{equation}
\mathcal{P}^{(l)} = \{ s \in \mathcal{S}^{(l)} \mid s > \tau^{(l)} \}.
\end{equation}
This ensures that subsequent prominence metrics are grounded in structurally meaningful transitions rather than incidental numerical fluctuations.

\paragraph{Quantifying Signal Prominence} 
Because reasoning transitions are more pronounced in specific functional layers, we introduce the \textbf{Spike Prominence Ratio (SPR)} to identify the optimal monitoring locus. 
Unlike a standard Signal-to-Noise Ratio (SNR), SPR utilizes median-based statistics on the set of detected candidates $\mathcal{P}^{(l)}$ to quantify how distinctly the reasoning pivots rise above the stochastic background:
\begin{equation}
\text{SPR}^{(l)} = \frac{\operatorname{median}(\mathcal{P}^{(l)}) - \operatorname{median}(\mathcal{S}^{(l)})}{\operatorname{MAD}(\mathcal{S}^{(l)})}
\end{equation}
This metric identifies the \emph{reasoning hub} within the model architecture where the signal for cognitive pivots is most reliable.

\paragraph{Spike Identification}
By selecting the representative layer $l^*$ that maximizes $\text{SPR}^{(l)}$, we obtain a temporally aligned indicator of high-magnitude latent transitions:
\begin{equation}
\mathcal{I}_t = \mathbbm{I}\left( \mathcal{S}^{(l)}_t > \tau^{(l)} \right)
\end{equation}
This indicator $\mathcal{I}_t$ serves as the foundational trigger for the STARS framework. 
It identifies the precise \emph{window of opportunity}—a state of elevated plasticity—enabling the subsequent diagnostic and steering phases to resolve the functional meaning of the transition in real-time. 
(Further details on layer selection and hyperparameter calibration are provided in Appendix~\ref{sec:AppendixB}).

\subsection{Stage 2: Select (Dynamics Diagnosis)}
\label{sec:phase_select}
The detection of a spike alone is insufficient for intervention, as it represents a fundamentally neutral state of plasticity. 
In this stage, STARS resolves this ambiguity by analyzing the geometry of the hidden-state trajectory to distinguish between benign transitions and pathological inertia.

\paragraph{Detecting Directional Flips}
We first test whether a spike corresponds to a conflict in reasoning direction by computing the cosine similarity between consecutive update vectors $v_t = h_t - h_{t-1}$ and $v_{t-1}$:
\begin{equation}
    c_t = \text{CosSim}(v_t, v_{t-1}).
\end{equation}
A sharp negative alignment ($c_t < -\tau_{\text{flip}}$) indicates a \textbf{Latent Flip}, reflecting a sudden reversal in the reasoning trajectory.

\paragraph{Diagnosing Recurrence}
For detected directional flips, we further assess whether the instability reflects a novel deviation or a recurrent state.
We maintain a memory bank $\mathcal{H}_{\text{bank}}$ that stores hidden states at previous flip events and define the \textbf{Recurrence Score} as:
\begin{equation}
    \rho_t = \max_{h_k \in \mathcal{H}_{\text{bank}}} \text{CosSim}(h_t, h_k).
\end{equation}
The memory bank is updated only at flip-aligned moments, ensuring that recurrence is evaluated relative to prior unstable states.
Based on $\rho_t$, we distinguish between:
\begin{itemize}[leftmargin=*,itemsep=1pt,topsep=2pt] 
    \item \textbf{Novel Instability ($\rho_t \le \tau_{\text{recur}}$):} The model is attempting a new direction, but with high conflict.
    \item \textbf{Cognitive Recurrence ($\rho_t > \tau_{\text{recur}}$):} The model has returned to a previous conflict state, indicating a reasoning loop.
\end{itemize}

\subsection{Stage 3: Steer (Adaptive Cueing)}
\label{sec:phase_modify}
Finally, STARS applies a lightweight intervention conditioned on the diagnosed dynamics. 
Rather than enforcing blunt termination, we inject state-aware natural language suffixes that prompt the model to recalibrate its path. 
These cues are structured to provide (i) \textit{state acknowledgment}, surfacing the diagnosed condition, and (ii) \textit{action guidance}, providing minimal bias for correction.

\paragraph{Shifting Suffix}
When a transition is identified as \textbf{Novel Instability}, we inject a \textbf{Shifting Suffix} (e.g., ``\textit{Wait. I am shifting my reasoning...}'') that explicitly acknowledges a sudden directional change in the reasoning trajectory and encourages the model to re-evaluate whether the new direction is grounded in the problem context.

\paragraph{Loop Breaker Suffix}
In contrast, when the state is diagnosed as \textbf{Cognitive Recurrence}, we inject a \textbf{Loop Breaker Suffix} (e.g., ``\textit{Wait. This line of thinking is looping...}'') that reflects the presence of a recurring reasoning state.
By surfacing the repetitive pattern, the suffix biases the model away from inertial continuation and toward exploring an alternative reasoning trajectory. We provide additional details on suffix design and thresholds in Appendix~\ref{sec:appendixD}.


\begin{table*}[t]
\caption{\textbf{Main Results.} Accuracy (\%) and average generation length (tokens) across benchmarks. 
Accuracy is reported as mean $\pm$ std over 5 runs. \textit{Abbreviations: M-Pert (MathPerturb), CondMath (ConditionedMath), Puzzle (PuzzleTrivial), GPQA-D (GPQA-Diamond).}
{\footnotesize
Bold and underlined entries indicate the best and second-best results, respectively.
}}
\vspace{-1mm}
\label{tab:all_results}
\centering
\footnotesize
\renewcommand{\arraystretch}{1.1}
\setlength{\tabcolsep}{2.7pt}

\begin{tabular}{l cc cc cc cc cc cc}
\toprule
\textbf{Method} &
\multicolumn{2}{c}{\textbf{M-Pert}} &
\multicolumn{2}{c}{\textbf{CondMath}} &
\multicolumn{2}{c}{\textbf{AIME24}} &
\multicolumn{2}{c}{\textbf{AIME25}} &
\multicolumn{2}{c}{\textbf{Puzzle}} &
\multicolumn{2}{c}{\textbf{GPQA-D}} \\
\cmidrule(lr){2-3}\cmidrule(lr){4-5}\cmidrule(lr){6-7}\cmidrule(lr){8-9}\cmidrule(lr){10-11}\cmidrule(lr){12-13}
& Acc. & Tok. & Acc. & Tok. & Acc. & Tok. & Acc. & Tok. & Acc. & Tok. & Acc. & Tok. \\
\midrule

\multicolumn{13}{l}{\cellcolor{gray!15}\textbf{DeepSeek-1.5B}} \\
Vanilla & \underline{62.08}\std{1.27} & 10038 & \underline{46.54}\std{4.28} & 11371 & \textbf{28.00}\std{4.52} & 17136 & 22.67\std{3.27} & 16192 & 28.48\std{6.53} & 8882 & 34.65\std{2.57} & 9842 \\
DEER    & 57.66\std{1.76} & 7019 & 46.15\std{5.96} & 8686  & \textbf{28.00}\std{2.67} & 14649 & 22.67\std{3.27} & 14922 & \underline{31.52}\std{4.11} & 14339 & \underline{35.15}\std{1.88} & 14591 \\
ConCISE & 58.70\std{3.62} & 7176 & \underline{46.54}\std{3.73} & 9302 & 26.67\std{2.98} & 11470 & \underline{25.33}\std{2.67} & 11900 & \underline{31.52}\std{4.54} & 9205 & 34.55\std{1.30} & 6953 \\
\textbf{STARS} & \textbf{62.60}\std{1.27} & 8571 & \textbf{53.08}\std{2.61} & 9018 & \textbf{28.00}\std{4.00} & 15075 & \textbf{27.33}\std{3.89} & 14708 & \textbf{37.58}\std{3.64} & 5289 & \textbf{35.45}\std{1.95} & 8032 \\
\midrule

\multicolumn{13}{l}{\cellcolor{gray!15}\textbf{DeepSeek-7B}} \\
Vanilla & \underline{82.34}\std{3.14} & 7104 & \underline{48.08}\std{5.01} & 9825  & \underline{53.33}\std{2.11} & 13010 & 39.33\std{1.33} & 14060 & 40.00\std{2.27} & 9646 & 47.47\std{1.72} & 7958 \\
DEER    & 74.03\std{1.16} & 6419 & 45.77\std{1.88} & 9692  & 49.33\std{6.80} & 14209 & \underline{40.67}\std{3.89} & 13456 & 37.58\std{5.62} & 6002 & 46.06\std{3.02} & 8840 \\
ConCISE & 81.04\std{2.80} & 4699 & 46.54\std{3.08} & 8499  & 47.33\std{6.46} & 9684  & 39.33\std{4.42} & 9866  & \underline{41.82}\std{4.02} & 8851 & \textbf{47.98}\std{2.28} & 5797 \\
\textbf{STARS} & \textbf{84.68}\std{0.97} & 6740 & \textbf{54.62}\std{1.96} & 8474 & \textbf{55.33}\std{3.40} & 12241 & \textbf{43.33}\std{2.98} & 14179 & \textbf{44.23}\std{3.09} & 5429 & \underline{47.78}\std{1.30} & 7863 \\
\midrule

\multicolumn{13}{l}{\cellcolor{gray!15}\textbf{Qwen3-4B}} \\
Vanilla & \underline{94.55}\std{2.08} & 12225 & 56.54\std{1.54} & 12630 & \underline{76.00}\std{3.89} & 19595 & 70.67\std{5.33} & 21064 & 73.33\std{3.53} & 4372 & \textbf{65.56}\std{1.25} & 9080 \\
DEER    & 90.65\std{1.72} & 9745  & \underline{56.92}\std{1.96} & 11252 & \underline{76.00}\std{4.90} & 17797 & 72.00\std{5.99} & 17936 & \underline{74.55}\std{3.64} & 4650 & 62.93\std{1.04} & 8766 \\
ConCISE & 93.77\std{1.91} & 5954  & 54.23\std{4.45} & 7444  & 75.33\std{3.40} & 8045  & \underline{74.67}\std{4.52} & 7240  & 72.73\std{2.71} & 3672 & 62.22\std{1.29} & 5181 \\
\textbf{STARS} & \textbf{95.32}\std{1.04} & 12931 & \textbf{58.85}\std{3.12} & 12924 & \textbf{77.33}\std{3.89} & 19403 & \textbf{77.33}\std{3.89} & 21069 & \textbf{76.36}\std{3.53} & 4556 & \underline{64.34}\std{1.62} & 9003 \\
\bottomrule
\end{tabular}
\vspace{-3mm}
\end{table*}

\section{Experiments}
\label{sec:experiments}

\subsection{Experimental Setup}

\paragraph{Datasets}
\label{sec:datasets}
We evaluate our framework on a collection of benchmarks covering complementary reasoning regimes.
Specifically, we use \textbf{ConditionedMath}~\citep{jang2025reasoning} and \textbf{MathPerturb}~\citep{huang2025mathperturb} to probe reasoning rigidity and perturbation-induced instability, and \textbf{AIME24} and \textbf{AIME25}~\citep{aime} to assess generalization under long-horizon mathematical reasoning.
In addition, we evaluate domain generalization using \textbf{PuzzleTrivial}\citep{jang2025reasoning}, which targets logical reasoning under disrupted intuitive priors, and \textbf{GPQA-Diamond}~\citep{rein2024gpqa}, which focuses on graduate-level scientific reasoning.
Detailed explanations are provided in Appendix~\ref{sec:appendixA_datasets}.

\paragraph{Models} 
We evaluate our framework on three open-source models spanning distinct scales and post-training paradigms.
We utilize \textbf{Qwen3-4B-Thinking-2507}~\citep{yang2025qwen3}, a compact reasoning-oriented model. 
Furthermore, to investigate the impact of post-training methodologies on latent dynamics, we employ \textbf{DeepSeek-R1-Distill-Qwen-1.5B}, which learns via distillation from a larger teacher, and \textbf{DeepSeek-R1-Distill-Qwen-7B}~\citep{guo2025deepseek}.

\paragraph{Baselines}
We compare our approach against representative inference-time intervention strategies from prior work.
All baselines considered in this comparison are \textbf{training-free} and operate purely at inference time, ensuring a controlled evaluation without additional model fine-tuning.
To address inference-time redundancy, we include \textbf{DEER}~\citep{yang2025dynamic}, which applies dynamic early-exit decisions based on token-level confidence signals.
We also evaluate the training-free decoding variant of \textbf{ConCISE}~\citep{qiao2025concise}, which guides generation using step-wise confidence signals. More details on baseline configurations are provided in Appendix~\ref{sec:appendixA}.

\paragraph{Hyperparameters}
All detection and diagnostic thresholds in STARS are calibrated on held-out samples disjoint from evaluation benchmarks. We use MAD to determine model-specific spike thresholds $\tau^{(l^*)}$ for noise resistance, while diagnostic thresholds ($\tau_{\text{flip}}, \tau_{\text{recur}}$) are derived from distributional analyses of cosine signals to isolate structural transitions. Once calibrated, these parameters remain fixed across all benchmarks to ensure generalization. See Appendices~\ref{sec:AppendixB} and \ref{sec:appendixD_threshold} for detailed protocols.

\subsection{Comparative Analysis across Reasoning Regimes}

Table~\ref{tab:all_results} summarizes the performance of STARS across all benchmarks.
Overall, STARS exhibits a favorable accuracy--efficiency trade-off across a broad range of tasks, with particularly pronounced benefits on benchmarks characterized by reasoning instability.

\paragraph{Mitigating Rigidity and Overthinking} 
On benchmarks such as \textbf{MathPerturb} and \textbf{ConditionedMath}, where failures are driven by redundant verification loops or local indecision, STARS often improves accuracy by effectively mitigating \textit{Cognitive Inertia}. 
For example, on ConditionedMath (DeepSeek-7B), STARS reduces the average generation length from 9,825 to 8,474 tokens while improving accuracy by 6.54\%. This indicates that spike-triggered steering can truncate unproductive cycles by redirecting the model away from inertial drift without suppressing necessary intermediate reasoning.

\paragraph{Sustaining Performance in Long-Horizon Reasoning}
In high-complexity settings like \textbf{AIME}, where extended reasoning is essential, STARS does not aggressively truncate inference. Instead, it selectively intervenes at detected \textbf{Critical Pivots}, preserving or slightly increasing the generation budget to maintain reasoning rigor. 
This reflects our design principle: rather than enforcing premature termination, STARS encourages exploration of alternative trajectories only when latent instability is observed.

\paragraph{Domain Generalization}
Beyond mathematics, STARS generalizes to non-mathematical domains. On \textbf{PuzzleTrivial}, it consistently improves accuracy by disrupting repetitive reasoning patterns born from strong intuitive priors. On \textbf{GPQA-Diamond}, STARS maintains performance comparable to vanilla baselines, ensuring that latent-based steering provides robust generalization across heterogeneous reasoning tasks without introducing regressions.

\paragraph{Error Analysis and Limitations}
Despite the consistent gains, we identify two primary failure modes of the STARS framework. First, in cases where the model terminates its reasoning before any salient latent spike occurs, no intervention can be triggered. Second, a detected spike may occasionally redirect the model toward an incorrect alternative trajectory, leading the steering cue to reinforce rather than rectify the underlying error. This typically occurs when the model's parametric priors are exceptionally dominant or when the initial reasoning path is fundamentally flawed beyond simple correction. Detailed qualitative examples and a full failure analysis are provided in Appendix~\ref{sec:appendixF}.

\subsection{Component Analysis and Ablations}
\label{sec:ablation}
To determine if active steering is essential or if simpler suppressive strategies suffice, we ablate STARS by isolating its modular components. This analysis validates the necessity of our three-stage cycle in overcoming the structural persistence of cognitive inertia.

\begin{table}[tb]
\caption{\textbf{Ablation Study on Intervention Mechanisms.} \textbf{Early Exit} evaluates halting versus continued reasoning, while \textbf{Flip-only} removes recurrence handling.}
\vspace{-1mm}
\label{tab:ablation_main}
\centering
\footnotesize
\renewcommand{\arraystretch}{1.1}
\setlength{\tabcolsep}{3.2pt}

\begin{tabular}{l cc cc}
\toprule
\textbf{Method} &
\multicolumn{2}{c}{\textbf{CondMath}} &
\multicolumn{2}{c}{\textbf{AIME25}} \\
\cmidrule(lr){2-3}\cmidrule(lr){4-5}
& Acc. & Tok. & Acc. & Tok. \\
\midrule

\multicolumn{5}{l}{\cellcolor{gray!15}\textbf{DeepSeek-1.5B}} \\
Vanilla    & 46.54\std{4.28} & 11371 & 22.67\std{3.27} & 16192 \\
Early Exit & 43.85\std{2.24} & 7250  & 22.67\std{3.89} & 13174 \\
Flip-only  & \underline{51.54}\std{3.92} & 9499  & \underline{23.33}\std{2.11} & 15387 \\
\textbf{STARS}
           & \textbf{53.08}\std{2.61} & 9017
           & \textbf{27.33}\std{3.89} & 14708 \\
\midrule

\multicolumn{5}{l}{\cellcolor{gray!15}\textbf{DeepSeek-7B}} \\
Vanilla    & 48.08\std{5.01} & 9825  & \underline{39.33}\std{1.33} & 14060 \\
Early Exit & 33.08\std{2.83} & 3457  & 28.67\std{4.52} & 9239 \\
Flip-only  & \underline{53.46}\std{4.62} & 8526  & 38.67\std{5.81} & 13820 \\
\textbf{STARS}
           & \textbf{54.62}\std{1.96} & 8473
           & \textbf{43.33}\std{2.98} & 12241 \\
\midrule

\multicolumn{5}{l}{\cellcolor{gray!15}\textbf{Qwen3-4B}} \\
Vanilla    & \underline{56.54}\std{1.54} & 12629 & 70.67\std{5.33} & 21064 \\
Early Exit & \underline{56.54}\std{2.88} & 9366  & 68.00\std{5.81} & 16953 \\
Flip-only  & 55.77\std{3.65} & 12740 & \underline{71.33}\std{3.40} & 21876 \\
\textbf{STARS}
           & \textbf{58.85}\std{3.12} & 12923
           & \textbf{77.33}\std{3.89} & 21069 \\
\bottomrule
\end{tabular}
\vspace{-6mm}
\end{table}

\paragraph{Steering vs. Simple Termination}
A straightforward alternative to steering is to simply terminate generation when a salient hidden-state spike is detected.
To examine this possibility, we evaluate a \textbf{Spike-based Early Exit} baseline, which halts generation at spike events when the model exhibits high next-token confidence, measured as the maximum softmax probability of the predicted token ($\mathcal{M}_t = \max_y p(y \mid h_t) > 0.9$).
As shown in Table~\ref{tab:ablation_main}, Early Exit substantially reduces generation length but leads to severe accuracy degradation across all models (e.g., a 15\% drop on DeepSeek-7B).
This failure mode suggests that high confidence at a spike often reflects \textit{Inertia of Direction}, where the model commits strongly to an internally consistent yet incorrect reasoning trajectory.
These results indicate that spike detection alone is insufficient, and that halting generation at such points discards potentially recoverable reasoning paths.

\paragraph{Impact of Recurrence Diagnosis}
To determine if directional correction alone is sufficient once an intervention is triggered, we evaluate a \textbf{Flip-only} variant of STARS, where steering is applied at detected directional flips while recurrence handling is disabled. As summarized in Table \ref{tab:ablation_main}, Flip-only STARS achieves a consistent but incomplete improvement over vanilla decoding across all models. While this variant successfully identifies and corrects abrupt directional conflicts, it consistently underperforms the full STARS model, particularly on benchmarks with longer reasoning horizons such as AIME25. This performance gap highlights the critical role of recurrence diagnosis in stabilizing the reasoning trajectory; by maintaining a memory of prior unstable states, STARS prevents the model from reverting to previously discarded inertial trajectories after an initial correction.

\paragraph{The Structural Synergy of Steering Cues}
\begin{table}[h]
\caption{\textbf{Suffix Structure Ablation}
on ConditionedMath using DeepSeek-7B.}
\label{tab:suffix_sensitivity}
\vspace{-1mm}
\centering
\footnotesize
\renewcommand{\arraystretch}{1.1}
\setlength{\tabcolsep}{3.2pt}

\begin{tabular}{lccc}
\toprule
\textbf{Suffix Variant} & \textbf{State} & \textbf{Action} & \textbf{Acc.} \\
\midrule
STARS (Full) & \checkmark & \checkmark & \textbf{54.62}\std{1.96} \\
State-only   & \checkmark & --          & 50.38\std{2.83} \\
Action-only  & --         & \checkmark  & 51.15\std{2.61} \\
Suffix-swap  & \checkmark & \checkmark  & \underline{53.85}\std{3.22} \\
\bottomrule
\end{tabular}
\vspace{-3mm}
\end{table}

Finally, we investigate whether the efficacy of STARS is determined solely by the \textit{timing} of the latent-triggered intervention or if the \textit{internal structure} of the steering cue is decisive. As detailed in Table \ref{tab:suffix_sensitivity}, we ablate the linguistic suffix into its two primary components: \textit{state acknowledgment} and \textit{action guidance}. The results indicate that removing either component degrades performance, demonstrating that neither element alone provides sufficient steering to disrupt the model's inertial trajectory. 
The \textit{suffix-swap} variant—injecting mismatched cues for a diagnosed state—attains intermediate performance. This suggests that while structural components are inherently beneficial at high-plasticity spike points, optimal steering requires \textbf{coherent semantic alignment} between the diagnosed state and its corresponding guidance. 
This reflects the flip-triggered nature of STARS, where even imperfectly aligned cues can provide a beneficial nudge near structurally meaningful transitions. However, peak accuracy is only achieved when the model’s internal state and the external cue are perfectly synchronized.

\subsection{Latent Dynamics and System Efficiency}
\label{sec:additional_analysis}

\begin{figure}[h]
    \centering
    \includegraphics[width=\linewidth]{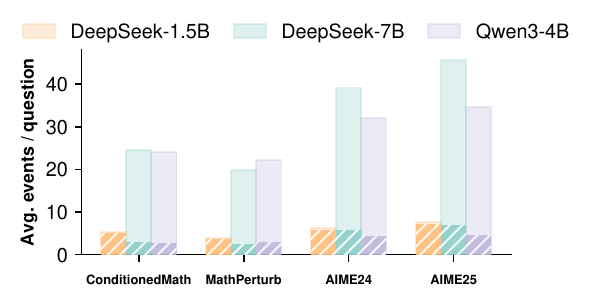}
    \vspace{-5mm}
    \caption{\textbf{Average spike and flip counts across model scales.} Semi-transparent bars indicate the average number of latent spikes detected per question, whereas solid bars denote directional flips, a structurally meaningful subset of spike events.
    }
    \label{fig:spike_flip_density}
    \vspace{-4mm}
\end{figure}

\paragraph{Scalability of Spike-Triggered Intervention} 
Figure~\ref{fig:spike_flip_density} shows how model scale influences the latent dynamics available for spike-triggered intervention across benchmarks.
We observe that larger models exhibit a substantially higher frequency of latent spike events, reflecting a richer space of candidate reasoning trajectories and more frequent high-plasticity transitions. Despite this increase in overall spike frequency, only a limited subset of spike events is identified as directional flips.
Importantly, the number of such flips remains relatively constrained across model scales, indicating that the proposed flip-based diagnosis selectively isolates structurally significant pivots rather than reacting to spike frequency alone.
The same pattern holds for the smaller DeepSeek-1.5B model: although spike activity is sparser due to a more compressed and less dynamically expressive reasoning process, flip-based diagnosis still identifies a small but meaningful subset of structurally significant transitions.
Together, these observations suggest that STARS operates as a robust selection mechanism, maintaining selective and stable intervention behavior even as the underlying latent dynamics vary substantially with model scale.

\begin{table}[h]
\centering
\caption{\textbf{Latency and throughput of baselines.}
Latency is measured as the mean end-to-end runtime per example,
and throughput as the average decoding speed (tokens/sec).}
\vspace{-1mm}
\label{tab:latency_throughput}
\footnotesize
\begin{tabular}{lcc}
\toprule
Method & Latency (s) & Throughput (tok/s) \\
\midrule
Baseline & 286.06 & 43.86 \\
DEER     & 416.92 & 35.72 \\
ConCISE  & 278.58 & 34.34 \\
STARS    & \textbf{251.34} & \textbf{47.09} \\
\bottomrule
\end{tabular}
\vspace{-4.5mm}
\end{table}

\paragraph{Inference Latency and Throughput} 
Table~\ref{tab:latency_throughput} compares the inference efficiency of STARS with prior training-free intervention baselines.
STARS achieves the lowest mean end-to-end latency while simultaneously attaining the highest decoding throughput among all compared methods.
This result is notable given that STARS performs inference-time steering, whereas competing approaches often rely on aggressive early termination or auxiliary confidence-based probes.
The efficiency of STARS arises from the lightweight nature of its intervention logic.
Flip and recurrence detection rely solely on hidden-state displacement signals and simple thresholding, introducing negligible computational overhead.
Moreover, by selectively correcting unproductive reasoning trajectories, STARS often reduces unnecessary generation length, leading to lower end-to-end latency than the vanilla baseline while preserving high decoding throughput.
As a result, the accuracy--efficiency gains of STARS stem from strategic trajectory correction rather than computationally expensive inference-time control.
A more detailed breakdown of latency sources and baseline behaviors is provided in Appendix~\ref{sec:appendixA_efficiency}.


\section{Related Works}
\subsection{Large Reasoning Models}
The paradigm of LLMs has evolved from scaling parameters to scaling test-time compute. 
Recent research has shown that allocating more inference budget—specifically by extending reasoning chains—can yield performance gains comparable to massive parameter scaling (e.g., s1~\citep{muennighoff2025s1}).
State-of-the-art models like o1~\citep{jaech2024openai} and DeepSeek-R1~\citep{guo2025deepseek} operationalize this principle, employing extended ``Chain of Thought'' (CoT) strategies to decompose intricate problems and perform self-correction. 
However, this test-time scaling introduces new challenges: as reasoning chains lengthen without guidance, models become prone to inefficiencies where the computational cost does not strictly correlate with performance gains, necessitating mechanisms to control and optimize the reasoning process.

\subsection{Layer-wise Dynamics and Latent Trajectory}
Prior works on interpretability have shifted from surface-level text analysis to probing internal representations. 
Prior work demonstrates that LLMs encode truthfulness and validity within their latent space, often decoupling internal knowledge from generated outputs~\citep{azaria2023internal, burns2023discovering}. 
This semantic information is not static but evolves hierarchically across layers.
\citet{chuang2024dola} highlighted this by contrasting distributions between early and late layers to isolate factual knowledge from noise. 
Building on these insights, we utilize layer-wise evolution not just for static probing, but to monitor dynamic reconfigurations—specifically identifying sudden state shifts as signals of abrupt internal transitions in reasoning.

\subsection{Inference-Time Intervention}
Prior work suggests that self-correction alone is often insufficient to override strong parametric priors, causing models to repeat familiar reasoning patterns or ignore counterfactual constraints~\citep{huanglarge, wu2024reasoning}.
Existing inference-time interventions broadly fall into two categories:
(i) activation-level steering that modifies internal representations~\citep{li2023inference, turner2024activation}, and
(ii) efficiency-driven methods that truncate generation using token-level heuristics such as entropy or confidence~\citep{yang2025dynamic, qiao2025concise}.
While effective at suppressing redundancy, these approaches either risk invasive interference or lack the ability to actively redirect an incorrect reasoning trajectory.
STARS addresses this gap by detecting latent semantic pivots and triggering lightweight, natural-language interventions to steer the model during internal instability.

\section{Conclusion}
In this paper, we introduced STARS, a training-free framework that mitigates cognitive inertia in LRMs by detecting Cognitive Pivots—functional boundaries marked by $L_2$ distance spikes in hidden states.
By employing a three-stage inference-time pipeline, STARS curtails redundant reasoning loops and counteracts rigid adherence to parametric priors.
Experiments across diverse benchmarks show that STARS achieves a superior accuracy-efficiency trade-off compared to existing token-level methods.


\section*{Impact Statement}
This work aims to improve the reliability and computational efficiency of large reasoning models by introducing a training-free mechanism for identifying and mitigating internal reasoning conflicts. 
By detecting moments of internal instability and selectively intervening during inference, the proposed approach reduces redundant reasoning behavior without suppressing necessary deliberation. 
From a sustainability perspective, limiting unnecessary inference paths can lower computational cost and energy consumption in large-scale deployments, contributing to more efficient use of test-time compute.

Beyond efficiency, mitigating reasoning rigidity may improve robustness in domains that require strict adherence to contextual or counterfactual constraints, such as scientific reasoning, mathematics, and technical decision-making. 
More broadly, our analysis highlights that internal hidden-state dynamics contain structured signals related to reasoning transitions, suggesting potential future directions for understanding model behavior, reasoning calibration, and interpretability.

At the same time, the effectiveness of spike-based signals depends on the interaction between model architecture and training strategy.
In particular, such signals may be less pronounced or harder to exploit in models with compressed representations, limited capacity, or training objectives that encourage smoother internal dynamics.

We do not foresee any specific negative societal consequences of this work beyond those generally associated with advances in machine learning, and we hope that the presented perspective encourages further responsible research into efficient and robust reasoning systems.

\bibliography{example_paper}
\bibliographystyle{icml2026}

\newpage
\appendix
\onecolumn

\section{Extended Experimental Setup}
\label{sec:appendixA}

\subsection{Dataset Specifications \& Filtering}
\label{sec:appendixA_datasets}

We evaluate STARS on a collection of reasoning benchmarks covering complementary failure modes of large reasoning models.

\textbf{ConditionedMath} and \textbf{PuzzleTrivial} are derived from the \textbf{ReasoningTrap} benchmark~\citep{jang2025reasoning}, which is designed to probe reasoning rigidity under counter-intuitive or disrupted constraints.
Within this framework, \textbf{ConditionedMath} modifies MATH- and AIME-style mathematical problems by introducing explicit counterfactual conditions, while \textbf{PuzzleTrivial} adapts well-known logic puzzles to invalidate intuitive default solution strategies.

For both datasets, we filter out instances with ambiguous or ill-defined ground truth, as well as near-duplicate problem structures, to ensure reliable evaluation.
After filtering, we retain 52 instances for \textbf{ConditionedMath} and 33 representative problems for \textbf{PuzzleTrivial}.

\textbf{MathPerturb}~\citep{huang2025mathperturb} is derived from Level-5 problems in MATH500 and targets perturbation-induced instability in mathematical reasoning.
We remove training-set overlaps and instances with undefined or non-unique answers, resulting in 77 evaluation samples.

All samples excluded during the filtering process for \textbf{ConditionedMath} and \textbf{MathPerturb} are not discarded, but are instead reserved exclusively for calibration.

To assess generalization under extended reasoning horizons, we additionally evaluate on \textbf{AIME24} and \textbf{AIME25}~\citep{aime}, which consist of held-out competition-style problems and are strictly disjoint from all calibration data.
Beyond mathematical reasoning, we further evaluate domain generalization on \textbf{GPQA-Diamond}~\citep{rein2024gpqa}, a high-difficulty subset of GPQA focusing on graduate-level scientific reasoning.

\subsection{Detailed Baseline Configurations}
\label{sec:appendixA_baseline}
We compare STARS against representative \textbf{training-free} inference-time baselines under identical decoding settings.
All methods operate without any additional fine-tuning, enabling controlled comparison on the same model checkpoints.

\textbf{DEER}~\citep{yang2025dynamic} is a training-free inference-time framework designed to reduce redundant reasoning by dynamically deciding when to terminate generation.
During decoding, DEER periodically induces a \emph{trial answer} from the current reasoning state and evaluates its reliability using token-level uncertainty signals.
If the induced answer exhibits sufficiently high confidence, generation is terminated early and the answer is returned; otherwise, decoding continues.
To mitigate the additional overhead introduced by trial answer induction and confidence evaluation, DEER further proposes a \emph{branch-parallel decoding} strategy.
This approach linearizes multiple candidate branches into a single sequence using specialized causal attention masks, allowing multiple trial answers to be evaluated within a single forward pass.
While effective at amortizing the cost of confidence probing, this mechanism requires maintaining multiple concurrent branches and incurs increased memory usage and implementation complexity.

In our experiments, we evaluate the standard single-trajectory configuration of DEER without branch-parallel acceleration.
This setting reflects a common deployment scenario where GPU memory constraints or system simplicity limit the feasibility of maintaining parallel reasoning branches.
Specifically, we apply entropy-based early stopping with a threshold of $0.95$, following the default inference-time configuration of the method.

\textbf{ConCISE}~\citep{qiao2025concise} performs confidence-guided step-wise compression by actively steering the generation process.
During decoding, ConCISE detects reflection steps and injects confidence phrases to regenerate the current step, effectively preventing low-utility reflective reasoning from being incorporated into the final chain.
In addition, it applies confidence-based early stopping once a sufficiently confident answer is produced, resulting in a compact final reasoning trace while preserving the standard answer format.

Importantly, while the final output produced by ConCISE is substantially shorter, this compression is achieved through regeneration rather than simple truncation.
Reflection steps are invalidated and re-generated with confidence injection, meaning that the total number of tokens generated internally during decoding can exceed the length of the final reasoning trace.

Although ConCISE is originally proposed as a training framework that combines confidence injection with early stopping, we adopt its \emph{training-free decoding variant} (ConCISE-Decoding) for a controlled inference-time comparison.
In this setting, confidence estimation and regeneration are applied directly during decoding, without relying on ground-truth verification or additional supervision.
We use step-wise confidence-guided decoding with an early-stopping threshold of $0.5$, following the commonly evaluated threshold regime reported in the original work.

\subsection{Implementation \& Hardware}
\label{sec:appendixA_imple}

\begin{figure}[h]
\centering
\begin{promptbox}{Default Prompt}
\texttt{Please reason step by step, and put your final answer within \textbackslash boxed\{\}.}
\end{promptbox}
\caption{The default prompt shared across all methods and benchmarks.}
\label{fig:default_prompt}
\end{figure}

All experiments are conducted using the HuggingFace transformers library.
We use the default decoding parameters (temperature and top-$p$) provided by each model to preserve their intended reasoning behavior.
The maximum sequence length is set to 32k tokens for all experiments.
Each reported result is averaged over 5 independent runs with different random seeds, and we report the mean accuracy along with the standard deviation.
All methods share the same default prompt shown in Figure~\ref{fig:default_prompt},
unless otherwise specified.
All experiments are performed on a single NVIDIA RTX 6000 Ada GPU.


\section{Detection Calibration and Sensitivity Analysis}
\label{sec:AppendixB}
This section details the calibration protocol used to establish the signal-driven detection parameters ($l^*, \tau^{(l^*)}$).
All calibration steps are conducted exclusively on held-out samples from ConditionedMath and MathPerturb, which are strictly disjoint from the test sets used for reporting evaluation results (Section~\ref{sec:appendixA_datasets}).
This separation ensures that calibration does not introduce information leakage into the final evaluation.

\subsection{Calibration Protocol}
\label{sec:AppendixB.1}
To ensure reliable spike detection across heterogeneous model architectures, we adopt a data-driven calibration pipeline that identifies the most stable and informative signal source, rather than fixing layers a priori.

\paragraph{Selection Procedure}
For each model, we perform a parameter sweep over a range of sensitivity multipliers $k \in \{5, 7,\dots, 13\}$ using a calibration dataset. The optimal configuration is selected based on a three-step hierarchy:
\begin{enumerate}
    \item \textbf{Density Filter:} We first discard any configuration that fails to meet the minimum density constraint (less than 1 event per 1k tokens), ensuring that selected signals reflect recurring reasoning dynamics rather than incidental numerical fluctuations.
    \item \textbf{Signal Maximization:} Among the valid candidates, we rank configurations by their \textbf{Spike Prominence Ratio (SPR)}.
    \item \textbf{Stability Check (Majority Voting):} We assess the stability of SPR rankings across independent trials via majority voting. This criterion is used as a secondary indicator of robustness rather than a hard selection constraint.
\end{enumerate}

\begin{table}[h]
    \centering
    \caption{\textbf{Calibrated Detection Parameters.} Comparing the primary (top row) and secondary (bottom row) configurations illustrates the stability of layer-wise signal selection across calibration trials.}
    \label{tab:calibrated_params}
    \small
    \begin{tabular}{l | c c c c}
        \toprule
        \textbf{Model} & \textbf{Layer} & \textbf{Thr ($\tau$)} & \textbf{SPR} & \textbf{Vote} \\
        \midrule
        \multirow{2}{*}{DeepSeek-1.5B} & L27 & 1.730 & 9.29 & 0.40 \\
                                       & L21 & 1.460 & 8.80 & 0.60 \\
        \midrule
        \multirow{2}{*}{DeepSeek-7B}   & L13 & 1.509 & 160.6 & 0.80 \\
                                       & L13 & 1.433 & 157.6 & 0.60 \\
        \midrule
        \multirow{2}{*}{Qwen3-4B} & L2 & 2.245 & 54.66 & 0.89 \\
                                           & L2 & 2.023 & 54.41 & 1.00 \\
        \bottomrule
    \end{tabular}
\end{table}

\subsection{Selected Configurations}
\label{sec:AppendixB.2}
Table~\ref{tab:calibrated_params} presents the final calibrated parameters. The results reveal distinct structural preferences based on model capacity.

For high-capacity models like \textbf{DeepSeek-7B} and \textbf{Qwen3-4B}, the selection converges robustly to a single layer with exceptionally distinct signals, indicating a centralized ``reasoning hub''.
In sharp contrast, the smaller \textbf{DeepSeek-1.5B} exhibits significantly weaker spike signals (SPR $<10$) and a lack of convergence in layer selection (the vote is split between Layer 27 and Layer 21). This suggests that in smaller models, the reasoning process is more distributed and less distinguishable from background generation noise.

\noindent Based on this analysis, we adopt the SPR-optimal configuration among density-valid candidates (corresponding to the first row for each model) as the fixed detection standard for all subsequent experiments.
When SPR ranking and vote stability disagree (e.g., DeepSeek-1.5B), we prioritize SPR while reporting the runner-up configuration for transparency.


\section{Qualitative Analysis of Latent Spikes}
\label{sec:appendixC}

To empirically validate the semantic implications of latent spikes proposed in Section~\ref{sec:semantic_analysis}, we conducted a granular qualitative analysis on the \textbf{ConditionedMath} benchmark, focusing on the three representative samples exhibiting the highest spike density for each model. 
Through manual annotation of 151 latent spikes isolated from these high-inertia trajectories, we observed that high-magnitude updates are predominantly \textit{Functional Pivots} (110 instances, 72.8\%), representing necessary operational shifts such as calculation initiation or symbolic derivation. 
In contrast, \textit{Critical Pivots}—markers of genuine inertia risk such as self-doubt or loop onset—constitute the minority (41 instances, 27.2\%). 

Crucially, as visualized in Figure~\ref{fig:pivot_analysis}, these two categories do not exhibit a clear separation when examined through text-aligned latent magnitudes alone. 
Despite their opposing semantic interpretations, both Functional and Critical Pivots often manifest as comparably high-magnitude updates in the hidden space.
This misalignment highlights a fundamental limitation of window-based, text-centric analyses: the semantic role of a pivot, as perceived by human annotators, does not necessarily coincide with the precise moment of internal state transition within the model. 
Consequently, interventions triggered purely by surface-level cues or post-hoc textual symptoms risk being mistimed, either suppressing valid reasoning steps or reacting after the trajectory has already diverged.
Motivated by this observation, STARS treats latent spikes not as errors to be corrected, but as temporal anchors for high-plasticity transitions, enabling subsequent trajectory-level diagnosis that operates directly on hidden-state dynamics rather than textual proxies.


\section{Methodological Sensitivity Analysis}
\label{sec:appendixD}

\subsection{Suffix Design}
\label{sec:appendixD_suffix}
The suffixes used in STARS are designed as lightweight, state-aware cues rather than corrective instructions.
They do not inject external knowledge or prescribe a specific solution path, but instead reflect the diagnosed reasoning state back to the model.

\begin{figure}[h]
\centering
\begin{promptbox}{Adaptive Steering Suffixes}
    \textbf{Shifting Suffix:} \\[0.5em]
    ``Wait. I am shifting my reasoning. Let's double-check if this direction is valid and grounded in the text.''

    \medskip

    \textbf{Loop Breaker Suffix:} \\[0.5em]
    ``Wait. This line of thinking is looping. Let's pause and pivot. Is there a simpler way to look at this problem?''
\end{promptbox}
\caption{The state-aware suffixes injected by STARS.}
\label{fig:steering_suffixes}
\end{figure}

Figure~\ref{fig:steering_suffixes} illustrates the two suffix types used in STARS.
The \textit{Shifting Suffix} is injected when a sudden directional change is detected, prompting the model to reassess whether the new trajectory is grounded in the problem context.
In contrast, the \textit{Loop Breaker Suffix} is triggered upon detecting a recurrent reasoning state, explicitly surfacing repetition and encouraging exploration of alternative trajectories.
Although both suffixes are expressed as short linguistic cues, they are differentiated by the underlying reasoning state they acknowledge.

To examine whether the effectiveness of these cues depends on their functional semantics rather than specific lexical forms, we decompose the suffix into two components: \emph{state acknowledgment} and \emph{action guidance}.
As shown in Table~\ref{tab:suffix_sensitivity}, removing either component degrades performance, while a misaligned \textit{suffix-swap} variant exhibits intermediate behavior.
This suggests that effective steering benefits from the presence of both components, with optimal performance achieved when state acknowledgment and action guidance are coherently aligned.

\subsection{Threshold Selection}
\label{sec:appendixD_threshold}

In this section, we provide empirical evidence supporting our choice of the flip threshold $\tau_{\text{flip}}$.
This analysis was conducted using the same held-out calibration set described in Appendix~\ref{sec:appendixC} to avoid data snooping on the evaluation benchmarks.
Rather than explicitly optimizing this parameter for performance, we analyze the distributional characteristics of flip signals and select a conservative operating point.

We quantify the separation between flipped and non-flipped updates using the standardized effect size $d'$:
\begin{equation}
d' = \frac{\mu_{\text{non-flip}} - \mu_{\text{flip}}}
{\sqrt{\frac{1}{2}(\sigma^2_{\text{non-flip}} + \sigma^2_{\text{flip}})}}
\end{equation}
where $\mu$ and $\sigma$ denote the mean and standard deviation of cosine similarities
within each group.

\begin{table}[h]
\caption{Directional flip statistics under varying $\tau_{\text{flip}}$.
$p_{\text{flip}}$ denotes the proportion of spikes classified as directional flips
($c_t < -\tau_{\text{flip}}$).}
\label{tab:flip_threshold_stats}
\centering
\begin{tabular}{c|cc}
\toprule
$\tau_{\text{flip}}$ & $p_{\text{flip}}$ & $d'$ \\
\midrule
0.05 & 0.65 & 1.97 \\
0.10 & 0.49 & 2.00 \\
0.15 & 0.34 & 2.08 \\
\textbf{0.20} & 0.22 & 2.20 \\
0.25 & 0.13 & 2.48 \\
0.30 & 0.08 & 3.12 \\
0.35 & 0.06 & 3.71 \\
0.40 & 0.05 & 4.26 \\
\bottomrule
\end{tabular}
\vspace{0.5em}
\end{table}

Table~\ref{tab:flip_threshold_stats} reports summary statistics of the flip signal collected at spike locations.
The table illustrates how the proportion of detected flips varies as a function of $\tau_{\text{flip}}$.

\begin{figure}[h]
\centering
\includegraphics[width=0.7\linewidth]{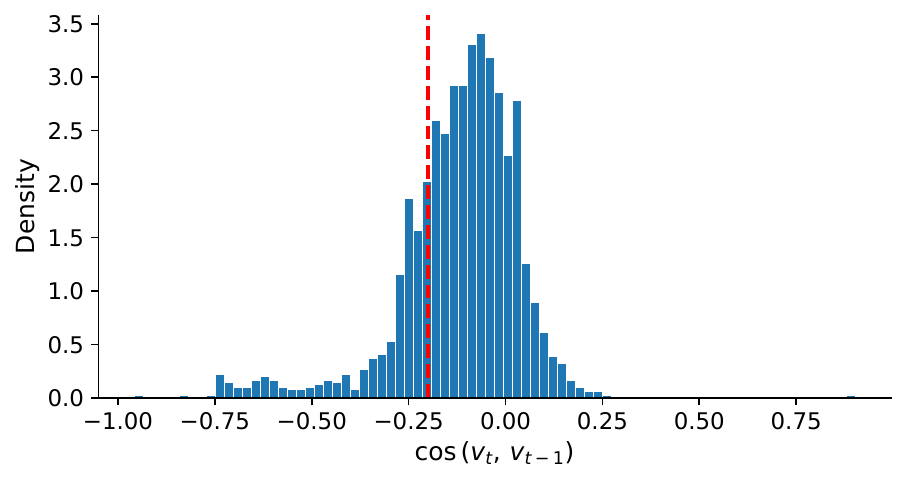}
\caption{Histogram of flip scores ($\cos(v_t, v_{t-1})$) at detected spike locations. The dashed red line indicates the selected flip threshold $-\tau_{\text{flip}}$ (here, $\tau_{\text{flip}}=0.2$), used to identify strong directional reversals.}
\label{fig:flip_histogram}
\end{figure}

Figure~\ref{fig:flip_histogram} visualizes the empirical distribution of flip scores.
The distribution is sharply concentrated around zero, with a substantial mass near small negative values.
This suggests that most spike-associated updates correspond to weak directional changes, while only a subset exhibit strong reversals beyond the operational flip threshold.

Based on this distribution, we set $\tau_{\text{flip}} = 0.2$.
As shown in Table~\ref{tab:flip_threshold_stats}, this value corresponds to a regime where the separation between flipped and non-flipped updates is already substantial ($d' \approx 2.2$),
while the proportion of detected flips remains moderate.
In particular, $\tau_{\text{flip}} = 0.2$ classifies roughly one-fifth of spike events as directional flips,
filtering out minor oscillations near zero cosine similarity without restricting detection to only extreme reversals.
This choice therefore represents a conservative operating point that balances discriminability and intervention frequency.

\begin{table}[h]
\caption{\textbf{Robustness Analysis over Flip and Recurrence Thresholds} using DeepSeek-7B.}
\label{tab:flip_recur_grid}
\centering
\footnotesize
\renewcommand{\arraystretch}{1.1}
\setlength{\tabcolsep}{3.0pt}

\begin{tabular}{c c cc}
\toprule
$\tau_{\text{flip}}$ & $\tau_{\text{recur}}$
& \textbf{CondMath} & \textbf{AIME24} \\
\midrule
\multirow{3}{*}{0.1}
 & 0.5 & 51.54\std{4.93} & 52.67\std{4.90} \\
 & 0.7 & 51.54\std{2.83} & 52.67\std{4.90} \\
 & 0.9 & 50.38\std{2.24} & 52.00\std{5.81} \\
\midrule
\multirow{3}{*}{0.2}
 & 0.5 & 54.48\std{4.53} & 50.67\std{5.73} \\
 & 0.7 & 53.08\std{3.12} & 51.33\std{4.00} \\
 & 0.9 & \textbf{54.62}\std{1.96} & \textbf{55.33}\std{3.40} \\
\midrule
\multirow{3}{*}{0.3}
 & 0.5 & 54.32\std{1.59} & 52.00\std{2.67} \\
 & 0.7 & 52.69\std{3.96} & 54.00\std{1.33} \\
 & 0.9 & 50.77\std{5.91} & 54.00\std{3.89} \\
\bottomrule
\end{tabular}
\end{table}

\subsection{Robustness of Threshold}
Finally, we analyze the robustness of STARS with respect to its steering thresholds, namely the flip threshold $\tau_{\text{flip}}$ and the recurrence threshold $\tau_{\text{recur}}$. 
Table~\ref{tab:flip_recur_grid} reports accuracy over a small grid of threshold combinations.

As shown in the table, performance varies non-trivially across threshold settings and different datasets favor different operating regimes.
Lower thresholds tend to trigger frequent interventions, which can improve responsiveness to cognitive inertia but risk destabilizing otherwise correct reasoning. 
Conversely, overly conservative thresholds reduce unnecessary interventions but may fail to correct persistent reasoning loops.

We therefore select the default configuration ($\tau_{\text{flip}}=0.2$, $\tau_{\text{recur}}=0.9$) as it achieves a favorable balance across datasets for DeepSeek-7B. 
This setting consistently mitigates cognitive inertia on ConditionedMath while simultaneously maintaining strong generalization performance on AIME24. 
Importantly, this choice is not optimal for any single benchmark in isolation, but represents the most stable trade-off observed under this model setting across distinct reasoning regimes.

Overall, these results indicate that while STARS is sensitive to threshold choices, its effectiveness does not depend on finely tuned parameters. 
Rather than prescribing a universal optimum, this analysis suggests that robust performance emerges from operating within a reasonable regime that balances inertia correction and generalization, which can be adapted across models if necessary.

\section{Extended Efficiency \& Latency Analysis}
\label{sec:appendixA_efficiency}

Latency and throughput are measured under a fixed evaluation protocol shared across all methods.
For each method, we report averages over multiple models and runs using identical hardware and decoding settings.

\begin{table}[h]
\centering
\caption{\textbf{Latency and throughput comparison (mean over models).}
Latency is measured as end-to-end runtime per example.
$p50$ and $p95$ denote the median and 95th-percentile latency, respectively.}
\label{tab:latency_full}
\footnotesize
\renewcommand{\arraystretch}{1.1}
\setlength{\tabcolsep}{4.2pt}
\begin{tabular}{lcccc}
\toprule
Method & Latency avg (s) & Latency p50 (s) & Latency p95 (s) & Throughput (tok/s) \\
\midrule
Baseline & 286.06 & 210.90 & 786.35 & 43.86 \\
DEER     & 416.92 & 239.89 & 1356.40 & 35.72 \\
ConCISE  & 278.58 & 210.99 & 724.91 & 34.34 \\
STARS    & \textbf{251.34} & \textbf{206.67} & \textbf{607.49} & \textbf{47.09} \\
\bottomrule
\end{tabular}
\end{table}

Table~\ref{tab:latency_full} provides a detailed latency and throughput breakdown across inference-time baselines.
DEER exhibits a substantially larger tail latency (p95), reflecting the input-dependent variability introduced by entropy-based early-exit decisions.
In particular, reasoning trajectories that include frequent self-reflection or explicit uncertainty markers tend to maintain high token-level entropy, delaying confident early-exit decisions and increasing runtime variance.
Because exit decisions are made dynamically based on intermediate confidence estimates, the amount of computation performed can vary widely across inputs, leading to long-tail latency behavior.
While DEER proposes branch-parallel decoding to amortize the cost of repeated confidence probing, this mechanism requires maintaining multiple concurrent reasoning branches and incurs additional memory overhead.
As a result, under single-trajectory decoding settings commonly used in practice, DEER can exhibit increased latency variance.

ConCISE reduces average latency in some cases by compressing low-utility intermediate steps, but incurs lower decoding throughput due to step-wise confidence evaluation and frequent intervention within the decoding loop.
Because this control is applied at the token or step level, decoding sequences that contain frequent self-reflective or corrective tokens (e.g., explicit reconsideration or revision cues) can trigger repeated confidence checks and compression decisions, further increasing control overhead.
This token-level control introduces additional synchronization and decision overhead, which can limit throughput despite reduced generation length.

It is important to note that the generation length reported for ConCISE reflects only the length of the final output trace, and does not account for the total number of tokens generated internally during decoding.
Due to its confidence-guided regeneration mechanism, ConCISE may repeatedly invalidate and re-generate intermediate steps when reflective reasoning is detected.
As a result, the internal decoding cost can substantially exceed the final generation length, despite the compact appearance of the output.
In our experiments, ConCISE generates on average approximately $1.6\times$ more tokens internally than those retained in the final trace, which helps explain why reduced output length does not necessarily translate into proportional latency or throughput gains.

In contrast, STARS shows consistently lower tail latency and stable median performance, indicating that its inference-time behavior remains predictable across inputs.
This stability arises from the sparse and event-driven nature of its intervention logic: flip and recurrence detection rely solely on lightweight hidden-state displacement signals and simple thresholding, and are triggered only at a small number of structurally significant points.
Moreover, by selectively correcting unproductive reasoning trajectories rather than performing continuous token-level control, STARS often reduces unnecessary generation length while preserving the standard autoregressive decoding flow.
As a result, STARS attains both lower end-to-end latency and higher decoding throughput than competing inference-time baselines.

These results suggest that the efficiency gains of STARS stem from stable and selective trajectory correction, rather than aggressive early termination or computationally intensive inference-time control mechanisms.


\section{Analysis of Explicit Instruction Prompting}
\label{sec:appendixE}

\begin{figure}[h]
\centering
\begin{promptbox}{Prompt Hinting Instruction}
\texttt{Please reason step by step, and put your final answer within \textbackslash boxed\{\}.\\ \textbf{An unusual condition is given. Do not overinterpret.}}
\end{promptbox}
\caption{The instruction used for Prompt Hinting.}
\label{fig:prompt_hinting}
\end{figure}

As a complementary point of comparison, we examine the effect of \emph{static prompt-based instruction}, which attempts to influence reasoning behavior solely through input-level linguistic cues.
As illustrated in Figure~\ref{fig:prompt_hinting}, Prompt Hinting prepends an explicit cautionary instruction designed to discourage overinterpretation, but applies no intervention once decoding begins.
This form of input-level guidance has been shown to yield measurable gains on benchmarks targeting \emph{reasoning rigidity}, where failures are often driven by early misinterpretation of counterfactual or unusual constraints~\citep{jang2025reasoning}.

A key limitation of this approach lies in its timing.
Because the instruction is provided only at the input stage, its influence is restricted to shaping the initial reasoning trajectory.
Once generation progresses and the model’s internal reasoning dynamics begin to drift under strong parametric priors, static instructions are frequently overridden, with no mechanism for mid-course correction.

\begin{table}[h]
\caption{\textbf{Effect of Prompt Hinting.} Comparison between explicit instruction prompting and dynamic steering on ConditionedMath and AIME25.}
\vspace{-2mm}
\label{tab:prompt_comparison}
\centering
\footnotesize
\renewcommand{\arraystretch}{1.1}
\setlength{\tabcolsep}{3.2pt}
\begin{tabular}{l cc cc}
\toprule
\textbf{Method} &
\multicolumn{2}{c}{\textbf{CondMath}} &
\multicolumn{2}{c}{\textbf{AIME25}} \\
\cmidrule(lr){2-3}\cmidrule(lr){4-5}
& Acc. & Tok. & Acc. & Tok. \\
\midrule

\multicolumn{5}{l}{\cellcolor{gray!15}\textbf{DeepSeek-1.5B}} \\
Vanilla & 46.54\std{4.28} & 11371 & 22.67\std{3.27} & 16192 \\
Prompt  & \underline{50.38}\std{4.77} & 10017 & \underline{23.33}\std{2.98} & 14947 \\
\textbf{STARS} & \textbf{53.08}\std{2.61} & 9018 & \textbf{27.33}\std{3.89} & 14708 \\
\midrule

\multicolumn{5}{l}{\cellcolor{gray!15}\textbf{DeepSeek-7B}} \\
Vanilla & \underline{48.08}\std{5.01} & 9825 & \underline{39.33}\std{1.33} & 14060 \\
Prompt  & 47.69\std{4.28} & 9913 & 38.00\std{1.63} & 13396 \\
\textbf{STARS} & \textbf{54.62}\std{1.96} & 8474 & \textbf{43.33}\std{2.98} & 12241 \\
\midrule

\multicolumn{5}{l}{\cellcolor{gray!15}\textbf{Qwen3-4B}} \\
Vanilla & 56.54\std{1.54} & 12630 & 70.67\std{5.33} & 21064 \\
Prompt  & \textbf{67.69}\std{4.11} & 11210 & \underline{76.00}\std{2.49} & 20751 \\
\textbf{STARS} & \underline{58.85}\std{3.12} & 12924 & \textbf{77.33}\std{3.89} & 21069 \\
\bottomrule
\end{tabular}

\vspace{-3mm}
\end{table}

Table~\ref{tab:prompt_comparison} reports a quantitative comparison between Prompt Hinting, vanilla decoding, and STARS on ConditionedMath and AIME25 across three model scales.
Prompt Hinting exhibits mixed behavior across settings.
Notably, on ConditionedMath with Qwen3-4B, static instruction yields a substantial accuracy improvement, suggesting that when reasoning failures primarily stem from early misinterpretation, pre-conditioning the model can be effective.
In this regime, the initial bias introduced by the instruction is sufficient to guide the model toward a more appropriate reasoning path.

However, this effect does not generalize consistently.
On larger models such as DeepSeek-7B, and on longer-horizon benchmarks such as AIME25, Prompt Hinting provides limited or unstable gains, and in some cases underperforms vanilla decoding.
These results indicate that static instruction alone is insufficient when reasoning errors arise from downstream instability, repetition, or inertia that emerges during generation.

Overall, this analysis highlights that explicit instruction can influence reasoning behavior, but its effectiveness is highly sensitive to model scale and task structure, and fundamentally constrained by its lack of temporal adaptivity.
This motivates the use of inference-time methods such as STARS, which intervene selectively at moments of detected instability rather than relying solely on input-level guidance.


\section{Qualitative Case Studies}
\label{sec:appendixF}
The qualitative examples in this section contrast vanilla decoding and STARS under identical settings.
In both cases, vanilla decoding falls into \textbf{Cognitive Inertia}, characterized by prolonged hesitation, repeated self-correction, and late-stage commitment to incorrect interpretations.
Concretely, the model repeatedly signals uncertainty through phrases such as “Maybe I’m misinterpreting the problem”, “Perhaps my initial assumption is wrong”, or “That seems too straightforward”, while continuing to elaborate on the same flawed hypothesis (Fig.~\ref{fig:vanilla_case_study_1} and Fig.~\ref{fig:vanilla_case_study_2}).
Rather than triggering early revision, these self-doubts accumulate into extended verification loops, eventually culminating in a confident but incorrect conclusion.
This process results in reasoning failure accompanied by substantial token overhead.

By contrast, STARS intervenes at detected pivot points and steers the trajectory back toward a coherent interpretation.
As illustrated in Fig.~\ref{fig:stars_case_study_1} and Fig.~\ref{fig:stars_case_study_2}, the injected suffixes explicitly prompt the model to reassess its current line of reasoning.
These interventions do not introduce new information, but instead encourage a structural re-evaluation of the existing reasoning, allowing the model to abandon unstable interpretations before they become entrenched.
As a result, STARS recovers correct solutions with significantly reduced token consumption, while preserving the underlying reasoning structure.

However, STARS is not universally effective.
We identify two recurring failure modes.
First, STARS may fail when the reasoning process terminates before any salient latent spike is observed.
In such cases, there is no opportunity for suffix injection or trajectory correction
(Fig.~\ref{fig:stars_failure_1}).
In such cases, the model converges prematurely on a solution before entering a sufficiently plastic internal state.
Second, even when a spike is detected and an intervention is triggered, the injected suffix can occasionally steer the model toward an incorrect alternative trajectory rather than resolving the underlying inconsistency
(Fig.~\ref{fig:stars_failure_2}).
Here, the model responds to the intervention by confidently reorganizing its reasoning around an erroneous assumption, leading to a different but still incorrect conclusion.

These cases highlight inherent limitations of spike-triggered intervention.
While latent spikes often coincide with meaningful internal transitions, they do not guarantee correct redirection.
Some forms of reasoning error may either precede detectable spikes or persist despite intervention, underscoring that STARS functions as a selective mitigation mechanism rather than a universal safeguard against reasoning failure.

\begin{figure*}[h]
\centering
\begin{promptbox}{Example 1: Vanilla}

\textbf{Question.}
\vspace{0.5em}

For each positive integer $n$, let $a_n$ be the least positive integer multiple of $23$ such that
$a_n \equiv 1 \pmod{7}$. Find the number of positive integers $n \le 1000$ such that
$a_n = a_{n+1}$.

\medskip
\textbf{Generated Text.}
\vspace{0.5em}

Okay, so I have this problem where for each positive integer $ n $, $ a_n $ is defined as the least positive integer multiple of 23 such that $ a_n \equiv 1 \pmod{7} $. I need to find how many positive integers $n$ less than or equal to 1000 satisfy $a_n = a_{n+1}$.

\vspace{0.5em}
\centerline{\dots}
\vspace{0.5em}

Therefore, $k \equiv 4 \pmod{7}$. So, the smallest positive integer $k$ is 4.
Therefore, $a_n = 23 \times 4 = 92$.
Wait, so is $a_n$ always 92? ... If $a_n$ is always 92, then $a_n = a_{n+1}$ for all $n$, so the number of such $n$ would just be 1000. \textbf{But that seems too straightforward.} The problem says ``positive integers $n$ less than or equal to 1000''

\vspace{0.5em}
\centerline{\dots}
\vspace{0.5em}

\textbf{Maybe I'm misinterpreting the problem.}
Wait, perhaps $a_n$ is defined for each $n$... \textbf{\textcolor{red}{Alternatively, perhaps it's the least multiple of 23 greater than or equal to $n$ that is congruent to 1 mod 7.}} That interpretation could make more sense, as it would vary with $n$. Let me try that interpretation.
... So, the sequence $a_n$ is 92, 253, 414, 575, 736, 897, 1058...

\vspace{0.5em}
\centerline{\dots}
\vspace{0.5em}

Therefore, in the sixth interval, $n=898$ to $n=999: 999 -898 +1=102$ numbers. 
Thus, total is 902. 
Therefore, the number of positive integers $n <=1000$ such that $a_n = a_{n+1}$ is 902.

\medskip
\textbf{Final Answer.} $\boxed{\textcolor{red}{902}}$ \hfill (9067 tokens)

\end{promptbox}
\caption{\textbf{Baseline failure on Example 1} (DeepSeek-7B on ConditionedMath).}
\label{fig:vanilla_case_study_1}
\end{figure*}


\begin{figure*}[h]
\centering
\begin{promptbox}{Example 1: STARS}

\textbf{Question.}
\vspace{0.5em}

For each positive integer $n$, let $a_n$ be the least positive integer multiple of $23$ such that
$a_n \equiv 1 \pmod{7}$. Find the number of positive integers $n \le 1000$ such that
$a_n = a_{n+1}$.

\medskip
\textbf{Generated Text.}
\vspace{0.5em}

Okay, so I need to figure out this problem where for each positive integer $ n $, $ a_n $ is the least positive integer multiple of 23 such that $ a_n \equiv 1 \pmod{7} $. Then, I have to find how many positive integers $n$ less than or equal to 1000 satisfy $a_n = a_{n+1}$.

Hmm, let me try to break this down step by step.

\vspace{0.5em}
\centerline{\dots}
\vspace{0.5em}

But that seems too straightforward, so maybe I need to check again. Wait, perhaps my confusion is because in the problem statement, ``the least positive integer multiple of 23 such that $a_n \equiv 1 \ \ \text{(mod}$

\vspace{0.5em}
\textbf{Wait. I am shifting my reasoning. Let's double-check if this direction is valid and grounded in the text.}
\vspace{0.5em}

The problem says: For each positive integer $n$, let $a_n$ be the least positive integer multiple of 23 such that $a_n \equiv 1 \pmod{7}$. So, each $a_n$ is a multiple of 23 such that $a_n \equiv 1 \pmod{7}$. So, indeed, this is a linear congruence...

\vspace{0.8em}
\centerline{\dots}
\vspace{0.8em}

But in the problem, it's written as $a_n \equiv 1 \pmod{7}$, without $n$, so it's fixed. Therefore, yeah, 92 is the same for all $n$.
Hence, the number of positive integers $n \le 10$

\vspace{0.5em}
\textbf{Wait. This line of thinking is looping. Let's pause and pivot. Is there a simpler way to look at this problem?}
\vspace{0.5em}

Maybe considering the definition of $a_n$. Each $a_n$ is the minimal multiple of 23 that is 1 mod7. Since the minimal solution is 92, which is fixed, so all $a_n = 92$.
Therefore, indeed, $a_n = a_{n+1}$ for all $n$, so every $n$ from 1 to 1000 satisfies it. So, the answer is 1000.
Thus, I think, despite my confusion, 1000 is the correct answer.

\medskip
\textbf{Final Answer.} $\boxed{1000}$ \hfill (4134 tokens)

\end{promptbox}
\caption{\textbf{STARS success on Example 1} (DeepSeek-7B on ConditionedMath).}
\label{fig:stars_case_study_1}
\end{figure*}


\begin{figure*}[h]
\centering
\begin{promptbox}{Example 2: Vanilla}

\textbf{Question.}
\vspace{0.5em}

There are $8!= 40320$ eight-digit positive integers that use each of the digits $1, 2, 3, 4, 5, 6, 7, 8$ exactly once. Let $N$ be the number of these integers that are divisible by $22$. Find the difference between $N$ and $2025$.

\medskip
\textbf{Generated Text.}
\vspace{0.5em}

Alright, so I have this problem here: I need to find how many eight-digit numbers, made by arranging the digits 1 through 8 exactly once each, are divisible by 22. Then, subtract 2025 from that number, $N$, and provide the result. Hmm, okay. Let me think about how to approach this.

\vspace{0.5em}
\centerline{\dots}
\vspace{0.5em}

But in each case, the sum is too high or too low.
\textbf{Wait, that can't be. So, perhaps my initial assumption is wrong.}
\dots
Therefore, in all cases, it's impossible to have sum of odd digit in even positions=2, since the smallest odd digit is 1. Therefore, there are no such numbers, so $N=0$. \textbf{But earlier, the combinatorial approach suggested N=128. So, which is correct?}

\vspace{0.5em}
\centerline{\dots}
\vspace{0.5em}

\textbf{\textcolor{red}{Therefore, $N=0$. Hence, $N -2025= -2025$.}} But perhaps the problem expects the absolute difference, which is 2025.
But since the problem says ``difference'', which is a signed number, so it's -2025. But since the problem is translated from Chinese, maybe the answer is 2025. But I think, given the analysis, the correct answer is $N=0$, so $N -2025= -2025$.
\vspace{0.5em}

But in the context of the problem, since $N=0$, the difference is -2025. Therefore, the final answer is -2025. But since the problem is presented in Chinese, and the translation is provided, maybe the answer is positive, but I think the correct answer is -2025.

\medskip
\textbf{Final Answer.} $\boxed{\textcolor{red}{-2025}}$ \hfill (18104 tokens)

\end{promptbox}
\caption{\textbf{Baseline failure on Example 2} (DeepSeek-7B on AIME 2025).}
\label{fig:vanilla_case_study_2}
\end{figure*}


\begin{figure*}[h]
\centering
\begin{promptbox}{Example 2: STARS}

\textbf{Question.}
\vspace{0.5em}

There are $8!= 40320$ eight-digit positive integers that use each of the digits $1, 2, 3, 4, 5, 6, 7, 8$ exactly once. Let $N$ be the number of these integers that are divisible by $22$. Find the difference between $N$ and $2025$.

\medskip
\textbf{Generated Text.}
\vspace{0.5em}

Okay, so I need to figure out how many eight-digit numbers, which use each of the digits 1 through 8 exactly once, are divisible by 22. Then, subtract 2025 from that number. Hmm, okay, let's break this down step by step.

\vspace{0.5em}
\centerline{\dots}
\vspace{0.5em}

Total subsets: $4 + 3 + 1 = 8$ subsets.
So, there are 8 subsets S of \{1,2,3,4,5,6,7,8\} with 4 elements that sum to 18.
But, since we have 8 such subsets, each contributing 2 * 3

\vspace{0.5em}
\textbf{Wait. I am shifting my reasoning. Let's double-check if this direction is valid and grounded in the text.}
\vspace{0.5em}

So, if each subset S corresponds to a unique set of digits in the even positions, which must include two even digits.

\vspace{0.5em}
\centerline{\dots}
\vspace{0.5em}

Therefore, for each subset S, the number of valid numbers is $2 * 3! * 4! = 2 * 6 * 24 = 288$.
Therefore, N = 8 * 2

\vspace{0.5em}
\textbf{Wait. This line of thinking is looping. Let's pause and pivot. Is there a simpler way to look at this problem?}
\vspace{0.5em}

Maybe I can approach it differently.
Wait, let me consider a smaller case to see. Suppose I have 4 digits, \{1,2,3,4\}, and I want numbers divisible by 6 (2 and 3).
Total N=4, which is correct. Thus, in the small case, method works.

\vspace{0.5em}
\centerline{\dots}
\vspace{0.5em}

Therefore, 2304 is indeed the correct count.
Hence, the difference $N - 2025 = 2304 - 2025 = 279$.
Thus, the answer is 279.

\medskip
\textbf{Final Answer.} $\boxed{279}$ \hfill (10240 tokens)

\end{promptbox}
\caption{\textbf{STARS success on Example 2} (DeepSeek-7B on AIME 2025).}
\label{fig:stars_case_study_2}
\end{figure*}


\begin{figure*}[h]
\centering
\begin{promptbox}{Failure Mode I: Termination before Suffix Injection}

\textbf{Question.}
\vspace{0.5em}

Sam is carrying a 2 gallon bucket of milk to his house from the barn, 3 miles away. The bucket leaks so that the amount of milk remaining is $(\frac{2}{3})^n$ gallons after $n$ miles. How many gallons of milk will be in the bucket when Sam gets home?

\medskip
\textbf{Generated Text.}
\vspace{0.5em}

Okay, so Sam has this 2-gallon bucket of milk, and he's walking from the barn to his house, which is 3 miles away. But here's the catch: his bucket is leaking. The problem says that after $n$ miles, the amount of milk left is $(2/3)^n$ gallons. Hmm, so I need to figure out how much milk is left when he gets home, which is after 3 miles.

\vspace{0.5em}
\centerline{\dots}
\vspace{0.5em}

After each mile, it's multiplying by $2/3$. So, after 1 mile: $2\cdot(2/3)=4/3$, but the formula says $(2/3)^1$ is $2/3$. Wait, now I'm confused.

Wait, let's test $n=0$. After 0 miles, it's $(2/3)^0 = 1$ gallon. 
Wait, but he starts with 2 gallons. So, the problem must be expressing the amount as $(2/3)^n$, so for $n=0$, it's 1, but he started with 2. 
That doesn't add up. Therefore, perhaps, the formula is missing a scaling factor.
\vspace{0.5em}
\textbf{\textcolor{red}{Therefore, perhaps the amount is $2\cdot(2/3)^n$.}}

\vspace{0.5em}
\centerline{\dots}
\vspace{0.5em}

Therefore, $M(n) = 2*(2/3)^n$. 
So, $M(3) = 2*(8/27) = 16/27$.

Alternatively, if the formula is just $(2/3)^n$, but that doesn't account for the initial 2 gallons, so I think that approach is wrong.

Therefore, the correct formula is $2*(2/3)^n$, so $16/27$ gallons.
So, in conclusion, after plugging in n=3, Sam will have $16/27$ gallons of milk when he gets home.

\medskip
\textbf{Final Answer.} $\boxed{\textcolor{red}{\frac{16}{27}}}$ \hfill (5112 tokens)

\end{promptbox}
\caption{\textbf{Failure Mode I: Termination before Suffix Injection} (DeepSeek-7B on ConditionedMath).}
\label{fig:stars_failure_1}
\end{figure*}


\begin{figure*}[h]
\centering
\begin{promptbox}{Failure Mode II: Misguided Suffix Steering}

\textbf{Question.}
\vspace{0.5em}

Among the 900 residents of Aimeville, there are 195 who own a diamond ring, 367 who own a set of golf clubs, and 562 who own a garden spade. In addition, each of the 900 residents owns a bag of candy hearts. Suppose the 234 residents who own exactly three of these things also own a candy store, which means all four items. Find the number of residents who own all four of these things.

\medskip
\textbf{Generated Text.}
\vspace{0.5em}

Okay, so I have this problem about residents in Aimeville and the different things they own. There are 900 residents in total. The things they can own are a diamond ring, a set of golf clubs, a garden spade, and a bag of candy hearts. The numbers given are: 195 own a diamond ring, 367 own a set of golf clubs, 562 own a garden spade, and all 900 own a bag of candy hearts.

\vspace{0.5em}
\centerline{\dots}
\vspace{0.5em}

In general, for each triplet of sets, the number of people in exactly that triplet is $|A \cap B \cap C| - |A \cap B \cap C \cap D|$.

\vspace{0.5em}
\textbf{Wait. I am shifting my reasoning. Let's double-check if this direction is valid and grounded in the text.}
\vspace{0.5em}

So, since the problem says that 234 residents own exactly three of these things and also own a candy store, meaning all four. So, in other words, exactly three of $A,B,C$, and all four because of $D$.
So, I think it's best to assume that these 234 residents actually own all four items.
So, then, the number of residents who own all four is 234.

\vspace{0.5em}
\centerline{\dots}
\vspace{0.5em}

\textbf{\textcolor{red}{But 0 = 672? That can't be. So, contradiction.}}
\textit{Hmm, that suggests that there's an inconsistency in the assumptions.}

\vspace{0.5em}
\centerline{\dots}
\vspace{0.5em}

Since $xyz = 234$, $10 + 2*234 = 10 + 468 = 478$, which matches $|\text{only }D| = 478$. 
So, all numbers check out.

Therefore, the number of residents who own all four is xyz = 234. Therefore, the answer is 234.

\medskip
\textbf{Final Answer.} $\boxed{\textcolor{red}{234}}$ \hfill (11225 tokens)

\end{promptbox}
\caption{\textbf{Failure Mode II: Misguided Suffix Steering} (DeepSeek-7B on ConditionedMath).}
\label{fig:stars_failure_2}
\end{figure*}


\end{document}